\documentclass[]{article}

\AtBeginDocument{%
  \providecommand\BibTeX{{%
    \normalfont B\kern-0.5em{\scshape i\kern-0.25em b}\kern-0.8em\TeX}}}

\usepackage{algorithm}
\usepackage{algpseudocode}
\usepackage{varwidth}
\usepackage{amsmath,epsfig}
\usepackage{subfig}
\usepackage{enumitem}
\usepackage{balance}
\usepackage{authblk}
\usepackage{url}
\usepackage{booktabs}

\graphicspath{{./images/}}
\DeclareGraphicsExtensions{.pdf,.jpeg, .png, jpg}

\newcommand{\hide}[1]{}

\begin{document}

\title{Image Retrieval using Multi-scale CNN Features Pooling}

\author{Federico Vaccaro, Marco Bertini, Tiberio Uricchio, Alberto Del Bimbo\\federico.vaccaro@stud.unifi.it, marco.bertini@unifi.it, tiberio.uricchio@unifi.it, alberto.delbimbo@unifi.it}
\affil{MICC - Università degli Studi di Firenze}

\maketitle

\begin{abstract}
In this paper, we address the problem of image retrieval by learning images representation based on the activations of a Convolutional Neural Network. We present an end-to-end trainable network architecture that exploits a novel multi-scale local pooling based on NetVLAD and a triplet mining procedure based on samples difficulty to obtain an effective image representation. Extensive experiments show that our approach is able to reach state-of-the-art results on three standard datasets.
\end{abstract}

\section{Introduction}\label{sec:intro}
Content-based image retrieval (CBIR) has received large attention from computer vision and multimedia scientific communities since the early 1990s. Texture, color and shape visual cues have been used to index images. For about 10 years, approaches based on local invariant features like SIFT and Bag-of-Words representations have obtained state-of-the-art results. Since the inception of Convolutional Neural Networks (CNNs), approaches using either convolutional or fully connected layer activations obtained better results \cite{razavian2014cnn} than those that aggregate local manually engineered features. The most recent CNN-based approaches aggregate regional activations, learning image representations in an end-to-end approach \cite{radenovic2018fine}.

In this paper, we present a novel multi-scale CNN regions pooling that aggregates local features before performing a second aggregation step using NetVLAD. This is used in an end-to-end learning approach in conjunction with a 3-stream Siamese network, to learn optimized image representations. A second contribution of this work is a triplet mining procedure that provides a diverse set of semi-hard and hard triplets, avoiding extremely hard ones that may hinder learning. The proposed method is evaluated on three standard image retrieval datasets: INRIA Holidays, Oxford5K and Paris6K, obtaining state-of-the-art results.

The paper is organized as follows: discussion of previous works is provided in Sect.~\ref{sec:previous}; description of the proposed method and its two contributions is given in Sect.~\ref{sec:method}; experiments on three standard CBIR datasets and a thorough comparison with competing state-of-the-art approaches are reported in Sect.~\ref{sec:experiments}; finally, conclusions are drawn in Sect.~\ref{sec:conclusion}.

\section{Previous work}\label{sec:previous}
Following the introduction of the Bag-of-Visual-Words model in \cite{sivic-2003}, many works have improved aspects such as approximating local descriptors \cite{Jegou-2010}, learning improved codebooks \cite{mikulik-2013}, improving local features aggregation \cite{perronnin-2010, jegou-2012, delhumeau2013revisiting}. However, following the success obtained using CNNs for image classification tasks, CNN-based features have started to be used also for image retrieval tasks. A thorough survey that compares SIFT and CNN-based approaches is provided in \cite{zheng2017sift}.

\subsection{CNN feature extraction}
The most straightforward approach is to use the activations of fully connected or convolutional layers as descriptors, using the networks as feature extractors. 
AlexNet FC6 has been used in \cite{razavian2014cnn}, outperforming local features approaches for instance retrieval in several standard datasets. In \cite{eccv2014-babenko} the performance of different AlexNet layers and the effects of PCA have been evaluated. More recent approaches use max-pooled activations from convolutional layers \cite{razavian-2014, azizpour-2014, zheng2016good}.

CNN features can be aggregated using techniques like Bag-of-Words, applied to local convolutional features as in \cite{mohedano2016bags}, VLAD, applied to global features as in \cite{yue2015exploiting} and to local patches as in \cite{gong2014multi, yue2015exploiting}, or using Fisher Vectors, e.g.~applied to localized local feature maps derived from objectness detectors as in \cite{iccv_vsm-2015}. Component-wise max-pooling of CNN features computed on object proposals has been used in \cite{mopuri-2015}. The approach used to compute CNN features in these methods may have an impact on the computational performance: the approaches used in \cite{gong2014multi, razavian2014cnn} require to compute CNN features on a large number of sub-patches, a problem that is reduced in \cite{iccv_vsm-2015, yue2015exploiting} where object proposals and ``dense sampling'' from max-pooling of convolutional layers are used. As a result, faster pooling approaches were introduced. Regional maximum activation of convolutions (R-MAC) aggregation \cite{tolias2016particular} consider a set of squared regions at different scales, and collects the maximum response in each channel. These descriptors are sum-pooled to create the final R-MAC descriptor. Hashing of CNN features, either global \cite{tmm-2017, morere-2017} or local, based on objectness scores \cite{xie-2015}, have been used to speed-up image retrieval tasks.

\subsection{End-to-end approaches}
In this class of methods CNN models are fine-tuned on a training set, learning better representations or aggregations of features, and allowing to extract features in an end-to-end manner through a single pass of the model. Typically this results in an improved performance w.r.t.~methods based on CNN feature extraction \cite{gordo2016deep, radenovic2016cnn}.

In \cite{arandjelovic2016netvlad} has been proposed a layer called NetVLAD, pluggable in any CNN architecture and trainable through back-propagation, that is inspired by the commonly used VLAD representation. This allows to train end-to-end a network, obtaining state-of-the-art results in image retrieval tasks using an aggregation of VGG16 convolutional activations.
Simultaneous learning of CNN and Fisher Vector parameters using a Siamese network and contrastive loss has been proposed in \cite{ong2017siamese}. 

Both the two current state-of-the-art approaches \cite{gordo2017end, radenovic2018fine} follow an end-to-end approach, one using a three-stream Siamese network with triplet loss and the other using a two-stream Siamese network with contrastive loss.

In \cite{gordo2017end} an end-to-end learned version of R-MAC descriptor is presented, along with a triplet mining procedure to efficiently train a three streams Siamese Network using triplet loss. In this approach, a region proposal network selects the most relevant regions of the image, where local features are extracted, in three scales of the input images.

In \cite{radenovic2018fine} a trainable Generalized-Mean (GeM) pooling layer is proposed, along with learning whitening, for short representations. Two stream Siamese network is trained using contrastive loss. The authors use structure-from-motion information and hard-matching examples for CNN training, and use up to 5 image scales to extract features.

Our proposed method shares similarity with all of these approaches, but in addition to our proposed pooling and triplet mining, it has important subtle differences that increase performance of the resulting system. Differently from \cite{gong2014multi, iccv_vsm-2015, yue2015exploiting, yue2015exploiting} our method is fully trainable end-to-end; differently from \cite{yue2015exploiting} multiple scales and only one convolutional layer are used; differently from \cite{gong2014multi} the VLAD aggregation is performed contemporarily at all the scales, and differently from \cite{iccv_vsm-2015} there is no use of region proposals. Differently from \cite{arandjelovic2016netvlad}, our input to the NetVLAD layer is not directly convolutional activations, but the concatenation of two max-pooled sets of activations.

\section{The Proposed Method}\label{sec:method}
The idea is to train a CNN network which provides optimized descriptors to perform image retrieval. The proposed method is inspired by the approaches used in \cite{arandjelovic2016netvlad, gordo2017end, radenovic2018fine}; the main differences are: \textit{i)} how the CNN features are collected using two different aggregation steps: the first one through max-pooling operations, i.e.~using 2-scales local features, followed by VLAD; \textit{ii)} the triplet mining procedure used to train a three-stream Siamese network, that selects semi-hard and hard triplets, avoiding those that could be considered as extremely hard, i.e.~whose visual similarity is very low due to minimal overlap, extreme zooming, etc. that may lead to overfitting and loss of generalization \cite{radenovic2018fine}.

\begin{figure}
  \includegraphics[width=0.9\textwidth]{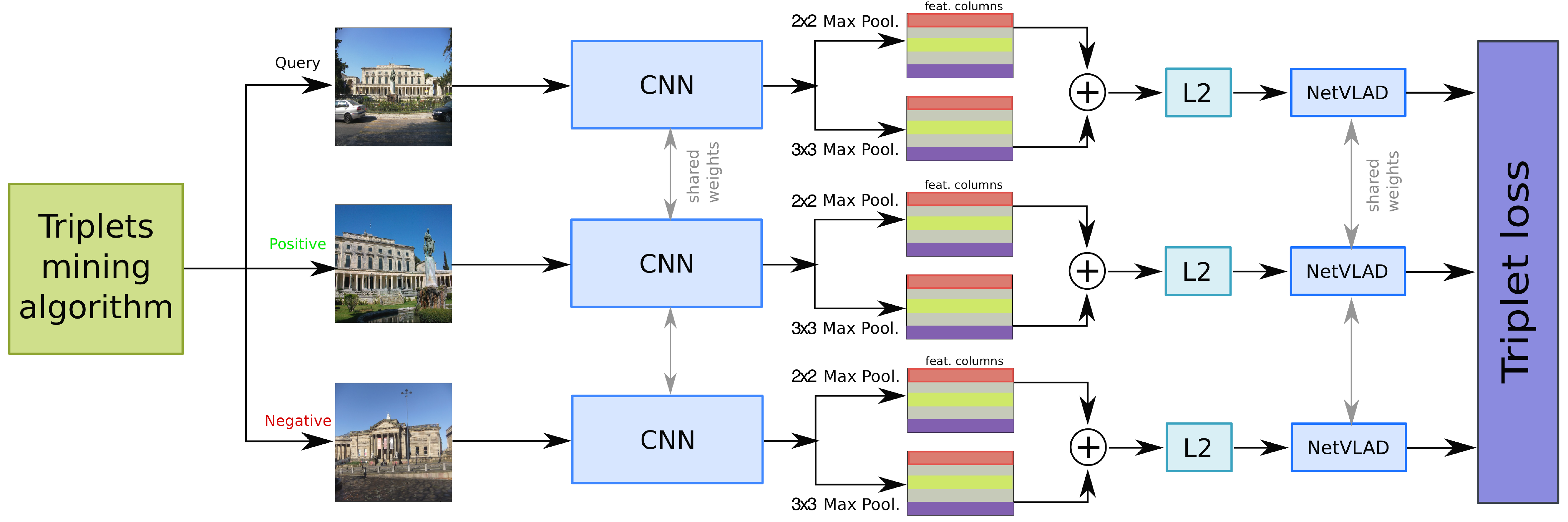}
  \caption{Schema of the proposed method: the three stream Siamese network is used at training time. At test time the query image is fed to the learned network to produce an effective image representation used to query the database.}
  \label{fig:teaser}
\end{figure}

\subsection{Pooling of local CNN features}\label{sec:pooling}
\begin{figure}[!htb]
  \centering
  \includegraphics[width=0.75\linewidth]{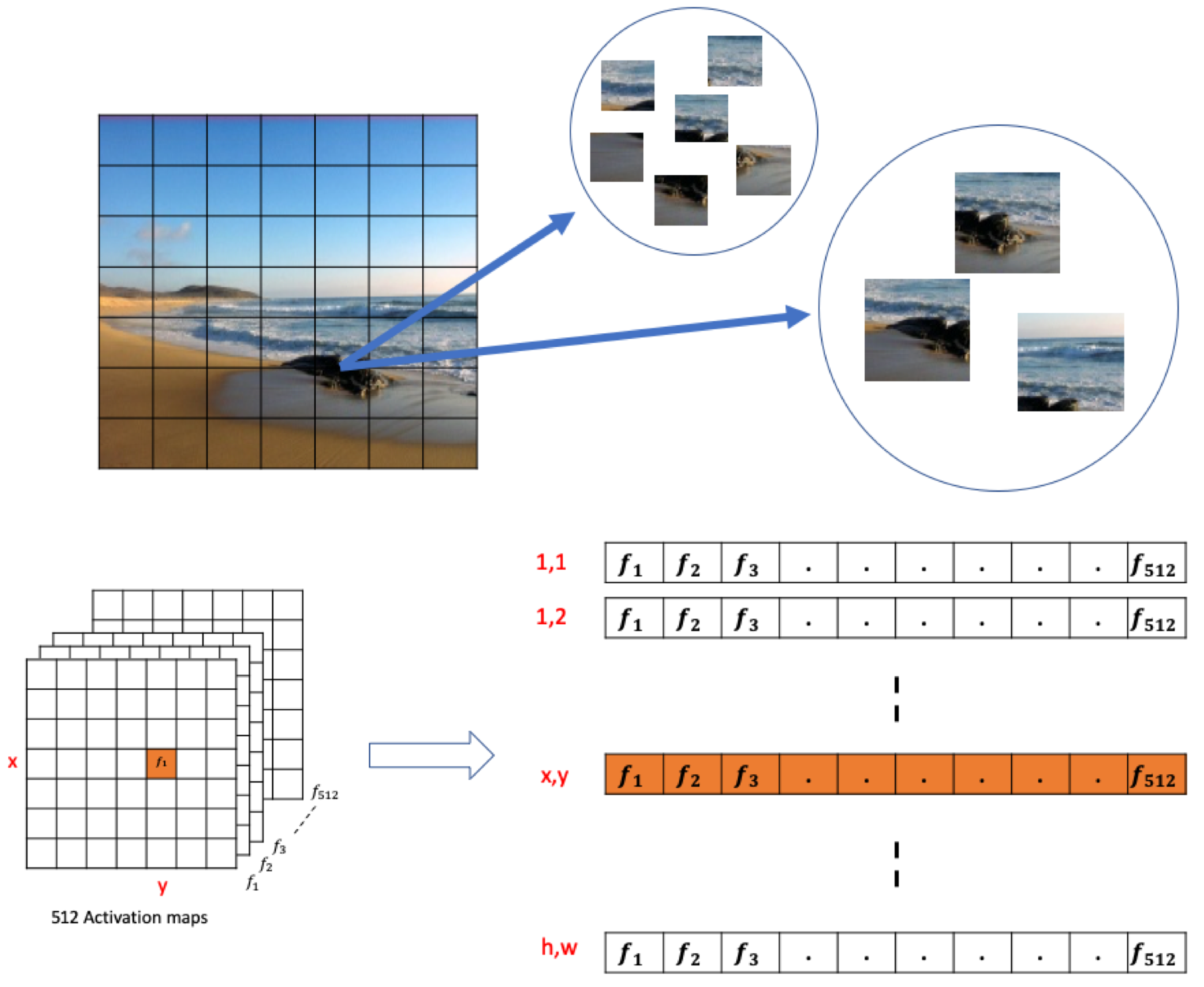}
  \caption{``Column feature'' extraction: \textit{top)} max-pooling with different scales, \textit{bottom)} activation maps collection as column features: this is performed at each pooling scale.}
  \label{fig:column-feature}
\end{figure}

Convolutional features are max-pooled using a $2 \times 2$ and $3 \times 3$ (both using stride=1) process, so to obtain representations at finer and larger detail. For each location of the two partitions the $f$ activation maps are collected, creating a $1 \times 1 \times f$ ``column feature'' (as defined in \cite{zheng2017sift}). This process, shown in Fig.\ref{fig:column-feature}, is akin to dense grid-based sampling of SIFT descriptors \cite{iscen2015comparison}. Sets of column features are concatenated, to provide a multi-scale descriptor of the image.

All the local CNN features are then aggregated using a NetVLAD \cite{arandjelovic2016netvlad} layer. The activations of this layer are used as a descriptor of the content of the image.
The NetVLAD layer is initialized with a K-Means clustering\footnote{In the experiments we performed it on MirFLICKR25K dataset \url{http://press.liacs.nl/mirflickr/mirdownload.html}}. As in \cite{arandjelovic2016netvlad} for NetVLAD we use $K=64$, resulting in a 32k-D representation.

\begin{figure*}[!htb]
  \centering
  \includegraphics[width=0.9\linewidth]{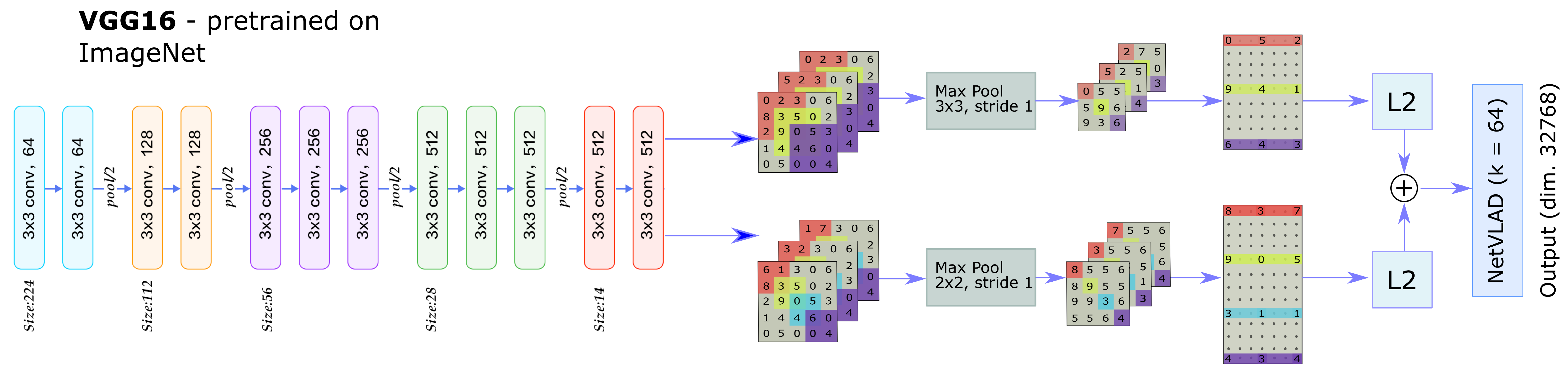}
  \caption{Overview of the proposed architecture, using VLAD aggregation of local multiscale max-pooling CNN features. VGG16 pre-trained on ImageNet is used as backbone.}  \label{fig:vgg16-vlad}
\end{figure*}

The approach can be applied in principle to any CNN network. In the following experiments we have tested VGG16, as it is commonly used in many competing methods and comparisons. An overview of the method is shown in Fig.~\ref{fig:vgg16-vlad}. The figure shows that we have used the penultimate convolutional layer in the $5^{th}$ block, since initial experiments have shown that using the last layer led to a reduced performance.

\subsection{Training and Triplet Mining}\label{sec:triplet-mining}
In this work we use a ranking loss based on triplets of images; the idea is to learn a descriptor so that the representation of relevant images is closer to the descriptor of the query than that of irrelevant ones. The design of the network is shown in Fig.~\ref{fig:teaser}: the weights of the convolutional layers of the CNN network and the NetVLAD layer are shared between the streams, since their size is independent of the size and aspect ratio of the images.

At training time the network is provided with image triplets. Given a query image $Q$ with descriptor $q$, a positive image $P$ with descriptor $p$, a negative image $N$ with descriptor $n$, a distance $d()$ (squared Euclidean distance) and a scalar $\alpha$ that controls the margin, the triplet loss used is $L = max(\alpha + d(q, p) - d(q, n), 0)$.
$\alpha$ is set to $0.1$ as in \cite{arandjelovic2016netvlad}.

An issue that may arise with this approach is due to the sampling of the triplets: e.g.~a random approach may select triplets that do not incur in any loss and thus do not improve the model.
We note that triplets may have different impact on the learning depending on the difficulty they pose. Some examples may be well separable if they are from different objects and may be easily learnt. In the contrary, similar but different objects may be challenging to be separated correctly. We may classify triplets as:\\
\textbf{easy triplets:} $d(q, p) < d(q, n) + \alpha < d(q, n)$ do not really improve the model;\\
\textbf{semi-hard triplets:} $d(q, p) < d(q, n)$ but $d(q, p) + \alpha > d(q, n)$ - these are more useful than easy triplets but may not add enough information;\\
\textbf{hard triplets:} $d(q, n) < d(q, p)$ - they produce a high loss.

The algorithm shown in Alg.~\ref{alg:euclid} generates semi-hard and hard triplets (with a 0.5 probability, line 14) with the following logic:\\
\textbf{case A:} searches the index $j$ for the first negative w.r.t.~query. If the index is not the first then the index of the positive sample is $j-1$ (line 15), resulting in a semi-hard triplet.\\
\textbf{case B:} otherwise, searches the index of the first positive after the first negative, resulting in a hard triplet (lines 18-19).\\
\textbf{case C:} this deals with extremely hard triplets, e.g.~due to strong changes in visual content like zoom or very different point of views of the same scene (see Fig.~\ref{fig:extreme-hard}), so that positives are very far from the query, i.e.~the index of the first positive is farther than a threshold $t$ (line 21). In this case triplets are discarded, since they may lead to overfitting or poor generalization.

The number of classes $k$ used in Alg.~\ref{alg:euclid} is $512$, the $mining\_batch\_size$ is $2048$. The procedure select the triplets so that they belong to different classes (line 28), yielding on average $250$ triplets and returned as mini batches composed by $24$.

\begin{minipage}[!hbt]{\columnwidth}
\begin{algorithm}[H]
\caption{Triplet mining}\label{alg:euclid}
\begin{algorithmic}[1]
\Procedure{Triplet mining}{$mining\_batch\_size, k, landmarks, t$}
\State $\textit{Pick \textbf{k} random landmarks}$
\State $\textbf{X}, y \gets \textit{pick \textbf{mining\_batch\_size} random images from}$
\State \hspace{2.5cm} $\textit{the selected landmarks and their labels}$
\State $\textbf{features} \gets model.extract\_features(\textbf{X})$
\State $\textit{\textbf{triplets}[]} \gets \textit{new list()}$
\For {$i \in [1, mining\_batch\_size] $}
\State $feature = \textbf{features}[i]$; $query\_label = y[i]$
\State $indices[] \gets \textit{Compute k-NN of feature}$
\State $q \gets i$; $p \gets null$; $n \gets null$
\For {$j \in [1, mining\_batch\_size]$}
\If{$label[j] \not= query\_label \textit{ and } n = null$}
\State $n \gets j$
\If{$j > 2 \textit{ with Probability 0.5}$}
\State{$p \gets j-1$} 
\State \textbf{break}
\EndIf
\ElsIf {$label[j] = query\_label \textit{ and } n \not= null$ }
\State{$p \gets j$}
\EndIf

\If{$p \ne null \textbf{ and } n \ne null \textbf{ and } p - n < t$}
\State $triplet \gets (X[q], X[p], X[n])$
\State $\textbf{triplets}.append(triplet)$
\State \textbf{break}
\EndIf
\EndFor
\EndFor
\State \textit{Keep just one triplet per class}
\State \textbf{return triplets}
\EndProcedure
\end{algorithmic}
\end{algorithm}
\end{minipage}
\\
\begin{minipage}[c]{0.9\columnwidth}
\centering
  \includegraphics[width=0.8\linewidth]{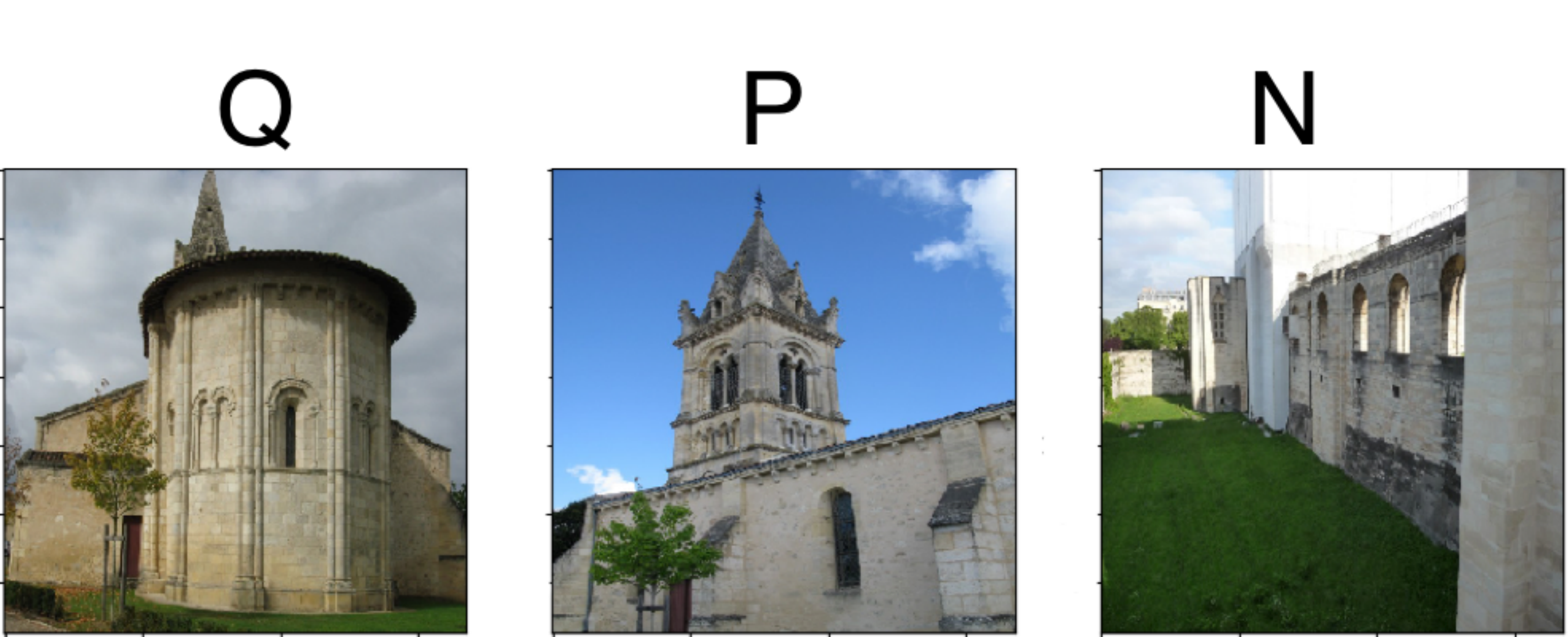}
  \captionof{figure}{Example of discarded extremely hard triplet.}
  \label{fig:extreme-hard}
\end{minipage}

\smallskip 
Training of the network is performed using Google Landmark V2 dataset\footnote{\url{https://github.com/cvdfoundation/google-landmark}}. In particular we use the train split of the ``cleaned'' version\footnote{\url{https://www.kaggle.com/confirm/cleaned-subsets-of-google-landmarks-v2}} presented in \cite{ozaki2019large}, that contains 1,580,470 images and 81,313 labels.
The mining process is performed every 8 iterations, to account for the fact that descriptors may change greatly, especially during the initial training. The network has been trained using the Adam \cite{adam2014} optimizer, with a starting learning rate of $10^{-5}$, decreased to $10^{-6}$ after few epochs. The training images have been resized to resolution $336 \times 336$, regardless to the original aspect-ratio.

\section{Experiments}\label{sec:experiments}
For the convolutional part of the network we evaluate a popular architecture, commonly used in other competing approaches, i.e.~VGG16, but other architectures can be plugged, as ResNet, etc.

\subsection{Datasets and Metrics}\label{sec:dataset}
We test our approach on three standard datasets: \textit{i)} Oxford5k dataset \cite{philbin2007object}, \textit{ii)} Paris6k dataset \cite{philbin-2008}, and \textit{iii)} INRIA Holidays dataset \cite{jegou-Holidays}; the standard evaluation protocol for these datasets is mean average precision (mAP). To be comparable with most CNN-based methods evaluations we manually correct the orientation of the images on the Holidays dataset, evaluating on the corrected images.

\subsection{Multi-scale Pooling and Image}
In the experiments reported in Tab.~\ref{tab:multi-pool}, we evaluate the effects of the first contribution of this work, i.e.~using two max-pooling to obtain multi-scale features before the NetVLAD layer. Results show that using both $2 \times 2$ and $3 \times 3$ pooling improve the performance. A single resolution image is used as input. It must be noted that all the results improve upon the standard NetVLAD pooling \cite{arandjelovic2016netvlad} reported in Tab.~\ref{tab:sota}, showing the benefit of the two-step local CNN features aggregation.

\begin{table}[!thb]
  \caption{Effects of multi-scale pooling (mAP).}
  \label{tab:multi-pool}
  \begin{tabular}{crrr}
    \toprule
    Pooling			&	Holidays			&	Oxford5k		&	Paris6k		\\
    \midrule
    $3 \times 3$	&	91.6				& 	81.0			&	87.3		\\
    $2 \times 2$	& 	88.8				& 	79.6			&	84.9		\\
    \textbf{Both}	& 	\textbf{92.3}		& 	\textbf{83.0}	&	\textbf{88.4}	\\
  \bottomrule
\end{tabular}
\end{table}

Different resolutions may provide different clues regarding the appearance of objects in the scene. Hence, we extract and combine features at different resolutions, improving the performance of the multi-scale pooling. In the experiments reported in Tab.~\ref{tab:multi-scale} we evaluate using different image resolutions at test time, evaluating the best combination on multiple datasets. Images are resized to $224\times224$, $336\times336$, $504\times504$ and $768\times768$ pixels, regardless of aspect ratio. The multi-resolution column reports the sizes used. In all these experiments multi-resolution pooling is used. Results show that image multi-resolution improves the performance. It is interesting to note that even the worst performing combination, i.e. without multi-resolution, the proposed method has better results than competing state-of-the-art approaches (see Tab.~\ref{tab:sota}).

\begin{table}[!thb]
  \caption{Effects of using multi-scale images, tested on INRIA Holidays (mAP).}
  \label{tab:multi-scale}
  \begin{tabular}{rrcc}
    \toprule
    Holidays	&	Oxford5k	&	Paris6k				&	Image resolutions	\\
    \midrule
    92.3 		&	83.0		&	88.4				&		336				\\
    93.2		& 	83.4		&	88.9				& 	336 + 504			\\
	\textbf{93.2}	&	\textbf{83.8}	& 	\textbf{89.3}		&	224 + 336 + 504		\\

	93.2		&	83.6	& 	89.3	&		224 + 336 + 504 + 768	\\
  \bottomrule
\end{tabular}
\end{table}

\subsection{Comparison with SOTA}\label{sec:sota}
In these experiments we evaluate the proposed method with current state-of-the art methods on all three datasets. Results are reported in Tab.~\ref{tab:sota}; all the methods reported in the table use VGG networks. Results of our method have been obtained using multi-resolution (224 + 336 + 504) and power normalization.

\begin{table}[!thb]
  \caption{Comparison with state-of-the-art methods (mAP).}
  \label{tab:sota}
  \begin{tabular}{lrrr}
    \toprule
    Method							& Holidays		&	Oxford5k			& 	Paris6k			\\
    \midrule
    \textbf{Our method}				& \textbf{93.2}	& 	83.8				&	\textbf{89.3}	\\
    GeM~\cite{radenovic2018fine}		& 89.5			&	\textbf{87.9}	&	87.7			\\
    R-MAC~\cite{gordo2017end}		& 89.1			&	83.1				&	87.1			\\	
    NetVLAD~\cite{arandjelovic2016netvlad}&	87.5		&	71.6				&	79.7			\\
    Fisher-Vector~\cite{ong2017siamese}&		-		&	81.5				&	82.4			\\
    BoW-CNN~\cite{mohedano2016bags}	&		-		&	73.9				&	82.0			\\
    R-MAC~\cite{tolias2016particular}&	86.9			&	66.9				&	83.0			\\
  \bottomrule
\end{tabular}
\end{table}

\section{Conclusions}\label{sec:conclusion}
We presented a novel multi-scale local CNN features pooling that, by exploiting end-to-end learning on a Siamese network, is able to learn an effective images representation. This is also thanks to a novel triplet mining procedure that is able to diversify triplets based on their difficulty and focus the learning on the most significative ones. Results on three standard datasets shows that the proposed approach obtains state-of-the-art results for the task of image retrieval.

\noindent \textbf{Acknowledgments.} We gratefully acknowledge the support of NVIDIA Corporation with the donation of the Titan X Pascal GPU used for this research.

\balance
\bibliographystyle{abbrv}
\bibliography{arxiv2020}

\begin{thebibliography}{10}

\bibitem{arandjelovic2016netvlad}
R.~Arandjelovic, P.~Gronat, A.~Torii, T.~Pajdla, and J.~Sivic.
\newblock Netvlad: Cnn architecture for weakly supervised place recognition.
\newblock In {\em Proc. of CVPR}, 2016.

\bibitem{azizpour-2014}
H.~{Azizpour}, A.~S. {Razavian}, J.~{Sullivan}, A.~{Maki}, and S.~{Carlsson}.
\newblock From generic to specific deep representations for visual recognition.
\newblock In {\em Proc. of CVPR Workshops}, June 2015.

\bibitem{eccv2014-babenko}
A.~Babenko, A.~Slesarev, A.~Chigorin, and V.~Lempitsky.
\newblock Neural codes for image retrieval.
\newblock In {\em Proc. of ECCV}, 2014.

\bibitem{delhumeau2013revisiting}
J.~Delhumeau, P.-H. Gosselin, H.~J{\'e}gou, and P.~P{\'e}rez.
\newblock Revisiting the vlad image representation.
\newblock In {\em Proc. of ACM MM}, 2013.

\bibitem{tmm-2017}
S.~Ercoli, M.~Bertini, and A.~Del~Bimbo.
\newblock Compact hash codes for efficient visual descriptors retrieval in
  large scale databases.
\newblock {\em IEEE Transactions on Multimedia (TMM)}, 19(11):2521--2532, Nov.
  2017.

\bibitem{gong2014multi}
Y.~Gong, L.~Wang, R.~Guo, and S.~Lazebnik.
\newblock Multi-scale orderless pooling of deep convolutional activation
  features.
\newblock In {\em Proc. of ECCV}, 2014.

\bibitem{gordo2016deep}
A.~Gordo, J.~Almaz{\'a}n, J.~Revaud, and D.~Larlus.
\newblock Deep image retrieval: Learning global representations for image
  search.
\newblock In {\em Proc. of ECCV}, 2016.

\bibitem{gordo2017end}
A.~Gordo, J.~Almazan, J.~Revaud, and D.~Larlus.
\newblock End-to-end learning of deep visual representations for image
  retrieval.
\newblock {\em International Journal of Computer Vision}, 124(2):237--254,
  2017.

\bibitem{iscen2015comparison}
A.~Iscen, G.~Tolias, P.-H. Gosselin, and H.~J{\'e}gou.
\newblock A comparison of dense region detectors for image search and
  fine-grained classification.
\newblock {\em IEEE Transactions on Image Processing}, 24(8):2369--2381, 2015.

\bibitem{jegou-Holidays}
H.~J\'egou, M.~Douze, and C.~Schmid.
\newblock Hamming embedding and weak geometric consistency for large scale
  image search.
\newblock In {\em Proc. of ECCV}, 2008.

\bibitem{Jegou-2010}
H.~J{\'e}gou, M.~Douze, and C.~Schmid.
\newblock Improving bag-of-features for large scale image search.
\newblock {\em International Journal of Computer Vision}, 87(3):316--336, 2010.

\bibitem{jegou-2012}
H.~{J{\'e}gou}, F.~{Perronnin}, M.~{Douze}, J.~{S{\'a}nchez}, P.~{P{\'e}rez},
  and C.~{Schmid}.
\newblock Aggregating local image descriptors into compact codes.
\newblock {\em IEEE Transactions on Pattern Analysis and Machine Intelligence},
  34(9):1704--1716, Sep. 2012.

\bibitem{adam2014}
D.~P. Kingma and J.~Ba.
\newblock Adam: A method for stochastic optimization.
\newblock In {\em Proc. of ICLR}, 2014.

\bibitem{mikulik-2013}
A.~Mikulik, M.~Perdoch, O.~Chum, and J.~Matas.
\newblock Learning vocabularies over a fine quantization.
\newblock {\em International Journal of Computer Vision}, 103(1):163--175,
  2013.

\bibitem{mohedano2016bags}
E.~Mohedano, K.~McGuinness, N.~E. O'Connor, A.~Salvador, F.~Marques, and
  X.~Giro-i Nieto.
\newblock Bags of local convolutional features for scalable instance search.
\newblock In {\em Proc. of ACM ICMR}, 2016.

\bibitem{morere-2017}
O.~Mor\`{e}re, J.~Lin, A.~Veillard, L.-Y. Duan, V.~Chandrasekhar, and
  T.~Poggio.
\newblock Nested invariance pooling and rbm hashing for image instance
  retrieval.
\newblock In {\em Proc. of ACM ICMR}, 2017.

\bibitem{ong2017siamese}
E.-J. Ong, S.~Husain, and M.~Bober.
\newblock Siamese network of deep {Fisher}-vector descriptors for image
  retrieval.
\newblock {\em arXiv preprint arXiv:1702.00338}, 2017.

\bibitem{ozaki2019large}
K.~Ozaki and S.~Yokoo.
\newblock Large-scale landmark retrieval/recognition under a noisy and diverse
  dataset.
\newblock {\em arXiv preprint arXiv:1906.04087}, 2019.

\bibitem{perronnin-2010}
F.~Perronnin, J.~S{\'a}nchez, and T.~Mensink.
\newblock Improving the {Fisher} kernel for large-scale image classification.
\newblock In {\em Proc. of ECCV}, 2010.

\bibitem{philbin2007object}
J.~Philbin, O.~Chum, M.~Isard, J.~Sivic, and A.~Zisserman.
\newblock Object retrieval with large vocabularies and fast spatial matching.
\newblock In {\em Proc. of CVPR}, 2007.

\bibitem{philbin-2008}
J.~{Philbin}, O.~{Chum}, M.~{Isard}, J.~{Sivic}, and A.~{Zisserman}.
\newblock Lost in quantization: Improving particular object retrieval in large
  scale image databases.
\newblock In {\em Proc. of CVPR}, June 2008.

\bibitem{radenovic2016cnn}
F.~Radenovi{\'c}, G.~Tolias, and O.~Chum.
\newblock Cnn image retrieval learns from bow: Unsupervised fine-tuning with
  hard examples.
\newblock In {\em Proc. of ECCV}, 2016.

\bibitem{radenovic2018fine}
F.~Radenovi{\'c}, G.~Tolias, and O.~Chum.
\newblock Fine-tuning cnn image retrieval with no human annotation.
\newblock {\em IEEE Transactions on Pattern Analysis and Machine Intelligence},
  41(7):1655--1668, 2018.

\bibitem{razavian-2014}
A.~Razavian, J.~Sullivan, A.~Maki, and S.~Carlsson.
\newblock A baseline for visual instance retrieval with deep convolutional
  networks.
\newblock {\em ITE Transactions on Media Technology and Applications}, 4, 12
  2014.

\bibitem{razavian2014cnn}
A.~S. Razavian, H.~Azizpour, J.~Sullivan, and S.~Carlsson.
\newblock {CNN} features off-the-shelf: An astounding baseline for visual
  recognition.
\newblock In {\em Proc. of CVPR Workshop of DeepVision}, 2014.

\bibitem{mopuri-2015}
K.~Reddy~Mopuri and R.~Venkatesh~Babu.
\newblock Object level deep feature pooling for compact image representation.
\newblock In {\em Proc. of CVPR Workshops}, June 2015.

\bibitem{sivic-2003}
J.~Sivic and A.~Zisserman.
\newblock Video google: a text retrieval approach to object matching in videos.
\newblock In {\em Proc. of ICCV}, Oct 2003.

\bibitem{tolias2016particular}
G.~Tolias, R.~Sicre, and H.~J{\'e}gou.
\newblock Particular object retrieval with integral max-pooling of cnn
  activations.
\newblock In {\em Proc. of ICLR}, 2016.

\bibitem{iccv_vsm-2015}
T.~Uricchio, M.~Bertini, L.~Seidenari, and A.~Del~Bimbo.
\newblock {Fisher} encoded convolutional {Bag-of-Windows} for efficient image
  retrieval and social image tagging.
\newblock In {\em Proc. of ICCV International Workshop on Web-Scale Vision and
  Social Media (VSM)}, 2015.

\bibitem{xie-2015}
L.~Xie, R.~Hong, B.~Zhang, and Q.~Tian.
\newblock Image classification and retrieval are one.
\newblock In {\em Proc. of ACM ICMR}, 2015.

\bibitem{yue2015exploiting}
J.~Yue-Hei~Ng, F.~Yang, and L.~S. Davis.
\newblock Exploiting local features from deep networks for image retrieval.
\newblock In {\em Proc. of CVPR Workshops}, 2015.

\bibitem{zheng2017sift}
L.~Zheng, Y.~Yang, and Q.~Tian.
\newblock Sift meets cnn: A decade survey of instance retrieval.
\newblock {\em IEEE Transactions on Pattern Analysis and Machine Intelligence},
  40(5):1224--1244, 2017.

\bibitem{zheng2016good}
L.~Zheng, Y.~Zhao, S.~Wang, J.~Wang, and Q.~Tian.
\newblock Good practice in cnn feature transfer.
\newblock {\em arXiv preprint arXiv:1604.00133}, 2016.

\end{thebibliography}

\end{document}